\documentclass[10pt, conference, compsocconf]{IEEEtran}
\ifCLASSINFOpdf
  \usepackage[pdftex]{graphicx}
  \usepackage{float}
  \usepackage[section]{placeins}
  \usepackage{float}
  \usepackage[export]{adjustbox}
  \usepackage{mathtools}
  \usepackage[T1]{fontenc}
  \usepackage[utf8]{inputenc}     
  \usepackage[english]{babel}   
  \usepackage{float}
  \usepackage{cite}
  \usepackage{hyperref}
\else
\fi

\begin{document}
\title{Object Detection Using Deep CNNs Trained on Synthetic Images}


\author{\IEEEauthorblockN{P. S. Rajpura}
\IEEEauthorblockA{Department of Electrical Engineering\\
Indian Institute of Technology\\
Gandhinagar, Gujarat, India 382355\\
Email:param.rajpura@iitgn.ac.in}
\and
\IEEEauthorblockN{H. Bojinov}
\IEEEauthorblockA{Innit Inc.\\
Redwood City, CA 94063,USA\\
Email:hristo.bojinov@innit.com}
\and
\IEEEauthorblockN{R. S. Hegde}
\IEEEauthorblockA{Department of Electrical Engineering\\
Indian Institute of Technology\\
Gandhinagar, Gujarat, India 382355\\
Email:hegder@iitgn.ac.in}

}


%Telephone: (800) 555--1212, Fax: (888) 555--1212}


\maketitle

\begin{abstract} 

  The need for large annotated image datasets for training Convolutional Neural
  Networks (CNNs) has been a significant impediment for their adoption in
  computer vision applications. We show that with transfer learning  an
  effective object detector can be trained almost entirely on synthetically
  rendered datasets. We apply this strategy for detecting packaged food products
  clustered in refrigerator scenes.  Our CNN trained only with 4000 synthetic
  images achieves mean average precision (mAP) of 24 on a test set with 55
  distinct products as objects of interest and 17 distractor objects.  A further
  increase of 12\% in the mAP is obtained by adding only 400 real images to
  these 4000 synthetic images in the training set. A high degree of photorealism
  in the synthetic images was not essential in achieving this performance.  We
  analyze factors like training data set size and 3D model dictionary size for
  their influence on detection performance. Additionally, training strategies
  like fine-tuning with selected layers and early stopping which affect transfer
  learning from synthetic scenes to real scenes are explored. Training CNNs with
  synthetic datasets is a novel application of high-performance computing and a
  promising approach for object detection applications in domains where there is
  a dearth of large annotated image data.  
  
\end{abstract}

% \fi
\IEEEpeerreviewmaketitle

\section{Introduction} \label{sec:intro}

The field of Computer Vision has reached new heights over the last few years. In
the past, methods like DPMs~\cite{Forsyth2014}, SIFT~\cite{Lowe2004} and
HOG~\cite{Dalal2005} were used for feature extraction, and linear classifiers were
used for making predictions. Other methods~\cite{Ekvall2003} used correspondences
between template images and the scene image. Later works focused on
class-independent object proposals~\cite{Uijlings2013} using segmentation and
classification using hand crafted features. Today methods based on Deep Neural
Networks (DNNs) have achieved state-of-the-art performance on image
classification, object detection, and segmentation~\cite{Krizhevsky2012,Szegedy2015}. DNNs been successfully
deployed in numerous domains~\cite{Krizhevsky2012,Szegedy2015}. Convolutional Neural
Networks (CNNs), specifically, have fulfilled the demand for a robust feature
extractor that can generalize to new types of scenes. CNNs were initially
deployed for image classification~\cite{Krizhevsky2012} and later extended to object
detection~\cite{Girshick2016}. The R-CNN approach~\cite{Girshick2016} used
object proposals and features from a pre-trained  object classifier. Recently
published works like Faster R-CNN~\cite{Ren2017} and SSD~\cite{Liu2016} learn object
proposals and object classification in an end-to-end fashion.

The availability of large sets of training images has been a prerequisite for
successfully training CNNs~\cite{Krizhevsky2012}. Manual annotation of images for object
detection, however, is a time-consuming and mechanical task; what is more, in some
applications the cost of capturing images with sufficient variety is
prohibitive. In fact the largest image datasets are built upon only a few categories
for which images can be feasibly curated (20 categories in PASCAL VOC~\cite{Everingham2015}, 80 in COCO~\cite{Lin}, and 200 in ImageNet~\cite{JiaDeng2009}). In applications where a large set of intra-category objects
need to be detected the option of supervised learning with CNNs is even
tougher as it is practically impossible to collect sufficient training
material.

\begin{figure*}[htbp]
\begin{center}
\includegraphics[width=0.7\textwidth]{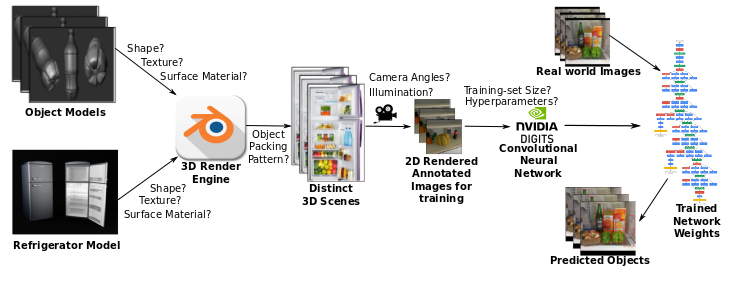}
\end{center}
   \caption{Overview of our approach to train object detectors for real images
   based on synthetic rendered images.}
\label{fig:Overview}
\end{figure*}

There have been solutions proposed to reduce annotation efforts by employing
transfer learning or simulating scenes to generate large image sets. The
research community has proposed multiple approaches for the problem of adapting
vision-based models trained in one domain to a different domain~\cite{Duan2012,Hoffman2013,Hoffman2014,Kulis2011,Long2015}. Examples include:  re-training a
model in the target domain~\cite{Yosinski2014}; adapting the weights of a
pre-trained model~\cite{Li2016}; using pre-trained weights for feature
extraction~\cite{Gupta2016}; and, learning common features between domains~\cite{Tzeng2014}.

Attempts to use synthetic data for training CNNs to adapt in real scenarios have
been made in the past. Peng et. al.\  used available 3D CAD models, both with and
without texture, and rendered images after varying the projections and
orientations of the objects, evaluating on 20 categories in the PASCAL VOC 2007
data set~\cite{Peng2014}. The CNN employed for their approach used a general
object proposal module~\cite{Girshick2016} which operated independently from the
fine-tuned classifier network. In contrast, Su and coworkers~\cite{Su2015} used
the rendered 2D images from 3D on varying backgrounds for pose estimation. Their
work also uses an object proposal stage and limits the objects of interest to a
few specific categories from the PASCAL VOC data set. Georgakis and coworkers~\cite{Georgakis2017} propose to learn object detection with synthetic data
generated by object instances being superimposed into real scenes at different
positions, scales, and illumination.  They propose the use of existing object
recognition data sets such as BigBird~\cite{Singh2014} rather than using 3D CAD
models. They limit their synthesized scenes to low-occlusion scenarios with 11
products in GMU-Kitchens data set.  Gupta et. al.\  generate a
synthetic training set by taking advantage of scene segmentation to create
synthetic training examples, however the goal is text localization instead of
object detection~\cite{Gupta2016}. Tobin et. al.\ perform domain randomization with
low-fidelity rendered images from 3D meshes, however their objective is to locate
simpler polygon-shaped objects restricted to a table top in world coordinates~\cite{Tobin2017}.
In~\cite{Ros2016,Handa2015}, the Unity game engine is used to generate RGB-D rendered images and semantic
labels for outdoor and indoor
scenes. They show that by using photo-realistic rendered images the effort for annotation can be significantly reduced. They combine synthetic and real data to train
models for semantic segmentation, however the network requires depth map
information for semantic segmentation. 

None of the existing approaches to training with synthetic data consider the use of synthetic image datasets
for training a general object detector in a scenario where high intra-class
variance is present along with high clutter or occlusion.  Additionally, while
previous works have compared the performance using benchmark datasets,  the
study of cues or hyper-parameters involved in transfer learning has not received
sufficient attention.  We propose to detect object candidates in the scene with
large intra-class variance compared to an approach of detecting objects for few
specific categories. We are especially interested in synthetic datasets which do
not require extensive effort towards achieving photorealism.  In this work, we
simulate scenes using 3D models and use the rendered RGB images to train a
CNN-based object detector. We automate the process of rendering and annotating
the 2D images with sufficient diversity to train the CNN end-to-end and use it
for object detection in real scenes. Our experiments also explore the effects of
different parameters like data set size and 3D model repository size.  We also
explore the effects of training strategies like fine-tuning selective layers and
early stopping~\cite{Yao2007} on transfer learning from simulation to reality.
The rest of this paper is organized as follows: our methodology is described in
\autoref{sec:method}, followed by the results we obtain reported in
\autoref{sec:results}, finally concluding the paper in \autoref{sec:conclusion}.
 
\section{Method} \label{sec:method}

Given a RGB image captured inside a refrigerator, our goal is to predict a
bound-box and the object class category for each object of interest. In
addition, there are few objects in the scene that need to be neglected. Our
approach is to train a deep CNN with synthetic rendered images from available 3D models.  Overview of the approach is shown in \autoref{fig:Overview}. Our work
can be divided into two major parts namely synthetic image rendering from 3D
models and transfer learning by fine-tuning the deep neural network with synthetic
images. 

\begin{figure*}[htbp]
\begin{center}
\includegraphics[width=0.75\textwidth]{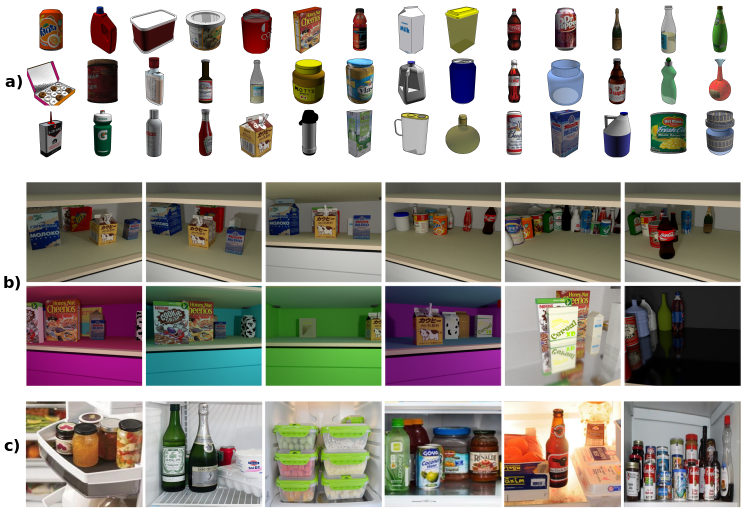}
\end{center}
   \caption{ Overview of the training dataset. a) Snapshots of the few 3D models from the ShapeNet database used for rendering images. We illustrate the variety in object textures, surface materials and shapes in 3D models used for rendering. b) Rendered non-photo realistic images with with varying object textures, surface materials and shapes arranged in random, grid and bin packed patterns finally captured from various camera angles with different illuminations. c) Few real images used to illustrate the difference in real and synthetic images. These images are subset of the real dataset used for benchmarking performance of model trained with synthetic images.  }
\label{fig:TrainSamples}
\end{figure*}

\subsection{Synthetic Generation of Images from 3D Models}

We use an open source 3D graphics software named Blender. Blender-Python APIs facilitate to load 3D models and automate the scene rendering.
We use Cycles Render Engine available with Blender since it supports ray-tracing to render synthetic images. 
Since all the required annotation data is available, we
use the KITTI~\cite{Geiger2012} format with bound-box co-ordinates,
truncation state and occlusion state for each object in the image.

Real world images have lot of information embedded about the environment, illumination, surface materials, shapes etc. Since the trained model, at test time must be able to generalize to the real world images, we take into
consideration the following aspects during generation of each scenario:

\begin{itemize}
  \item Number of objects
  \item Shape, Texture, and Materials of the objects
  \item Texture and Materials of the refrigerator
  \item Packing pattern of the objects
  \item Position, Orientation of camera
  \item Illumination via light sources
\end{itemize}

In order to simulate the scenario, we need 3D models, their texture information
and metadata. Thousands of 3D CAD models are available online. We choose
ShapeNet~\cite{Chang2015} database since it provides a large variety of objects of
interest for our application. Among various categories from ShapeNet like
bottles, tins, cans and food items, we selectively add 616 various object models
to object repository ($R_0$) for generating scenes. \autoref{fig:TrainSamples}a
shows few of the models in $R_0$. The variety helps randomize the aspect of
shape, texture and materials of the objects. For the refrigerator, we choose a
model from Archive3D~\cite{3D2015} suitable for the application. The design
of refrigerator remains same for all the scenarios though the textures and material
properties are dynamically chosen.

For generating training set with rendered images, the 3D scenes need to be
distinct. The refrigerator model with 5-25 randomly selected objects from $R_0$
are imported in each scene. To simulate the cluster of objects packed in
refrigerator like real world scenarios, we use three patterns namely grid,
random and bin packing for 3D models. The grid places the objects in a
particular scene on a refrigerator tray top at predefined distances. Random
placements drop the objects at random locations on refrigerator tray top. Bin
packing tries to optimize the usage of tray top area placing objects very close
and clustered in the scene to replicate common scenarios in refrigerator. The
light sources are placed such that illumination is varied in every scene and the images are not biased to a well lit environment since refrigerators
generally tend to have dim lighting. Multiple cameras are placed at random
location and orientation to render images from each scene.  The refrigerator texture and
material properties are dynamically chosen for every rendered image.
\autoref{fig:TrainSamples}b shows few rendered images used as training set while \autoref{fig:TrainSamples}c shows the subset of real world images used in training.

\begin{figure*}[htbp]
\begin{center}
\includegraphics[width=0.9\textwidth]{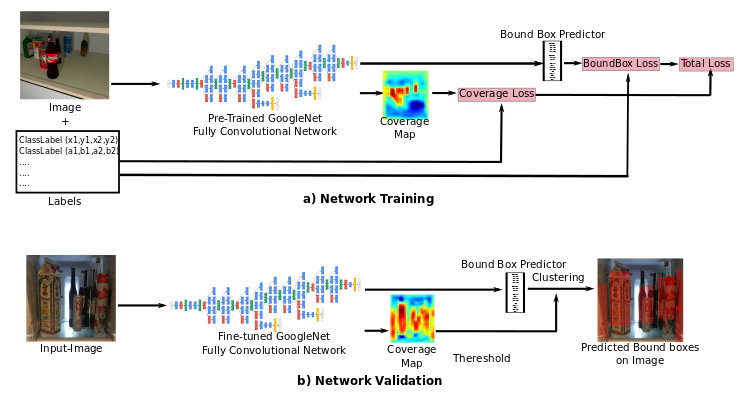}
\end{center}
   \caption{Work-flow for the major steps of the system. a) Using annotated
   images, the FCN generates a coverage map and bound-box co-ordinates. The
 training loss is a weighted sum of coverage and bound-box loss. b) At validation
 time, coverage map and bound boxes are generated from the FCN. }
\label{fig:TrainOverview}
\end{figure*}

\subsection{Deep Neural Network Architecture, Training and Evaluation}

\begin{figure}[htbp]
\begin{center}
{\includegraphics[width=0.9\linewidth]
{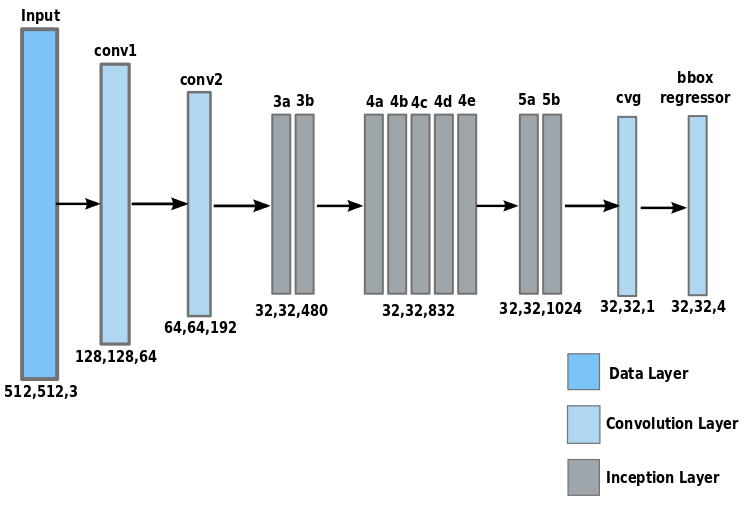}}
\end{center}
   \caption{The Fully Convolutional Network architecture used in the detector. Each bar represents a layer. Convolution layer includes Convolution, ReLU activation and Pooling. Inception Layer includes the module as described in GoogleNet~\cite{Szegedy2015}.  }
\label{fig:FCN}
\end{figure}

\autoref{fig:TrainOverview} provides the detailed illustration of network
architecture and work-flow for the training and validation stages. For neural
network training we use NVIDIA-DIGITS$^{TM}$-DetectNet~\cite{Barker2016} with
Caffe~\cite{Jia2014} library in back-end.  During training, the RGB images with
resolution (in pixels) 512 x 512 are labelled with standard KITTI~\cite{Geiger2012}
format for object detection. We neglect objects truncated or highly occluded in
the images using appropriate flags in the ground truth label generated while
rendering. The dataset is later fed into a fully convolutional network (FCN)
predicting coverage map for each detected class. The FCN network represented
concisely in \autoref{fig:FCN} has the same structure as
GoogLeNet~\cite{Szegedy2015} without the data input layers and output layers.
For our experiments, we use pre-trained weights on ImageNet to initialize the
FCN network which has earlier been helpful for transfer
learning~\cite{Georgakis2017}.

The bound-box regressor predicts bound-box corner per grid square. We train the detector through stochastic gradient descent with Adam optimizer using standard learning rate of $1e^{-3}$.
The total loss is the weighted summation of the following losses:
\begin{itemize}
  \item L2 loss between the coverage map estimated by the network and ground truth
  \begin{equation}
    \frac{1}{2N} \sum_{i=1}^{N}\big|coverage_i^t - coverage_i^p\big|^2
  \end{equation}
  where $coverage^t$  is the coverage map extracted from annotated ground truth
  and $coverage^p$ is the predicted coverage map while $N$ denoting the batch
  size.
  \item L1 loss between the true and predicted corners of the bounding box for the object covered by each grid square.
  \begin{equation}
      \frac{1}{2N} \sum_{i=1}^{N}\Big[\big|x_1^t - x_1^p\big|+\big|y_1^t - y_1^p\big|+\big|x_2^t - x_2^p\big|+\big|y_2^t - y_2^p\big|\Big]
  \end{equation}
  where $(x_1^t,y_1^t,x_2^t,y_2^t)$  are the ground-truth bound box co-ordinates
  while $(x_1^p,y_1^p,x_2^p,y_2^p)$ are the predicted bound box co-ordinates.
  $N$ denotes the batch size.
\end{itemize}

For the validation stage, we threshold the coverage map obtained
after forward pass through the FCN network, and use the bound-box regressor to
predict the corners. Since multiple bound-boxes are generated, we finally
cluster them to refine the predictions.  For evaluation, we compute Intersection
over Union (IoU) score. With a threshold hyper-parameter, predicted bound boxes
are classified as True Positives (TP), False Positives (FP) and False
Negatives (FN). Precision (PR) and Recall (RE) are calculated using these metrics
and a simplified mAP score is defined by the product of PR and RE~\cite{Hoiem2012}.

\section{Results and Discussion} \label{sec:results}

We evaluate our object detector trained exclusively with synthetically rendered
images using manually annotated crowd-sourced refrigerator images. \autoref{fig:PredictedInferences}
illustrates the variety in object textures, shapes, scene illumination and
environment cues present in the test set. The real scenarios also include other objects like vegetables,
fruits, etc.\ which need to be neglected by the detector. We address them as
distractor objects.  

All the experiments were carried on workstation with Intel$^R$ Core$^{TM}$ i7-5960X processor accelerated by NVIDIA$^{R}$ GEFORCE$^{TM}$ GTX 1070. NVIDIA-DIGITS$^{TM}$ (v5.0) tool was used to
prepare and manage the databases and trained models. Hyper-parameters search on
learning rate, learning rate policy, training epochs, batch-size were performed
for training all neural network models. 

The purpose of our experiments was to evaluate the efficacy of transfer learning
from rendered 3D models on real refrigerator scenarios. Hence we divide this section into two parts:
\begin{itemize}
  \item Factors affecting Transfer Learning: Here, we analyze the factors which we experimented with to achieve the best detection performance via transfer learning. We study following factors affecting overall detection performance:
\begin{itemize}
  \item Training Dataset Size: The variety in training images used  determines the performance of neural networks.
  \item Selected Layer Fine-tuning: Features learned at each layer in CNNs have been distinct and found to be general across domains and modalities. Fine-tuning of the final fully-connected linear classification layers has been used in practice for transfer learning across applications. Hence, we extend this idea to train several convolutional as well as linear layers of the network and evaluate the resulting performance.
  \item Object Dictionary Size: The appearance of an object in image in static environment is a function of its shape, texture and surface material property. Variance in objects used for rendering has been observed to increase detection performance significantly ~\cite{Su2015}.
\end{itemize}
  \item Detection Accuracy: Here, we represent the analysis of the performance on real dataset achieved with the best detector model\footnote{Trained network weights and synthetic dataset are available at \url{https://github.com/paramrajpura/Syn2Real}}.
\end{itemize}

\subsection{Factors affecting transfer learning}
\begin{figure}[htbp]
\begin{center}
{\includegraphics[width=0.9\linewidth]
{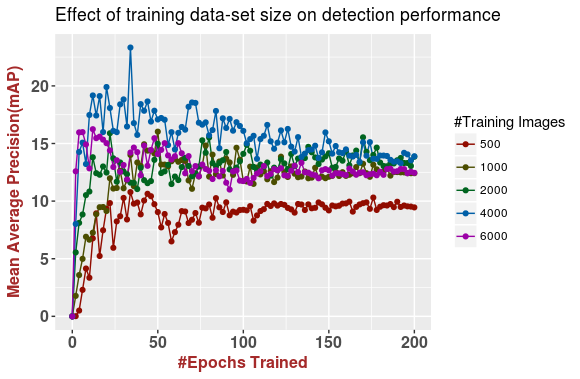}}
\end{center}
   \caption{Detection results for the validation image-set for the training iterations.}
\label{fig:TrainSizePlot}
\end{figure}

Considering other parameters like object dictionary size and fine-tuned network layers, we vary the training data size from 500-6000. We observe an increase in mAP up to 4000 images followed by a light decline in performance as shown in  \autoref{fig:TrainSizePlot}. Note that the smaller dataset is a subset of the larger dataset size i.e. we have incrementally added new images to train dataset. After an extent, we observe decline in accuracy as we increase the dataset size suggesting over-fitting to synthetic data with increase in dataset size.

\begin{figure}[htbp]
\begin{center}
{\includegraphics[width=0.9\linewidth]
{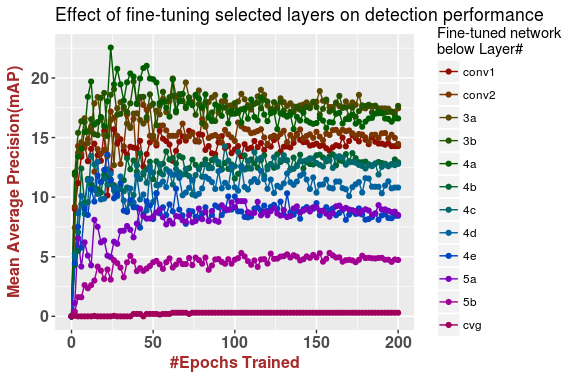}}
\end{center}
   \caption{Neurons in layers of CNN learn distinct features. Network weights were fine-tuned by freezing layers sequentially. The figure represents the performance with weights fine-tuned till mentioned layers.}
\label{fig:LayerWisePlot}
\end{figure}

\begin{figure}[htbp]
\begin{center}
{\includegraphics[width=0.9\linewidth]
{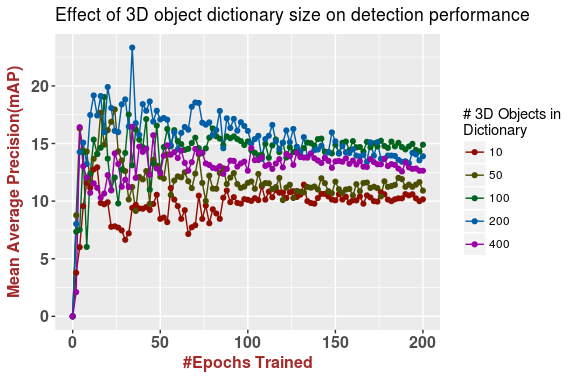}}
\end{center}
   \caption{Variety in training data affects the capability of generalizing object detection in real scenarios. }
\label{fig:DictSizePlot}
\end{figure}

We use GoogleNet FCN architecture with 11 different hierarchical levels with few
inception modules as single level (\autoref{fig:FCN}). mAP vs. number of epochs
chart is presented in \autoref{fig:LayerWisePlot} for models with different
layers selected for fine-tuning. Starting from training just the final coverage
and bounding-box regressor layers we sequentially open deeper layers for
fine-tuning. We observe that fine-tuning all the inception modules helps
transfer learning from synthetic images to real images in our application. The
results show that selection of the layers to fine-tune proves to be important
for detection performance. 

To study the relationship of variance in 3D models with performance, we
incrementally add distinct 3D models to the dictionary starting from 10 to 400.
We observe an increase in mAP up to 200 models and slight decline later on as
represented in \autoref{fig:DictSizePlot}.

\subsection{Detection Accuracy}
\begin{figure}[htbp]
\begin{center}
{\includegraphics[width=\linewidth]
{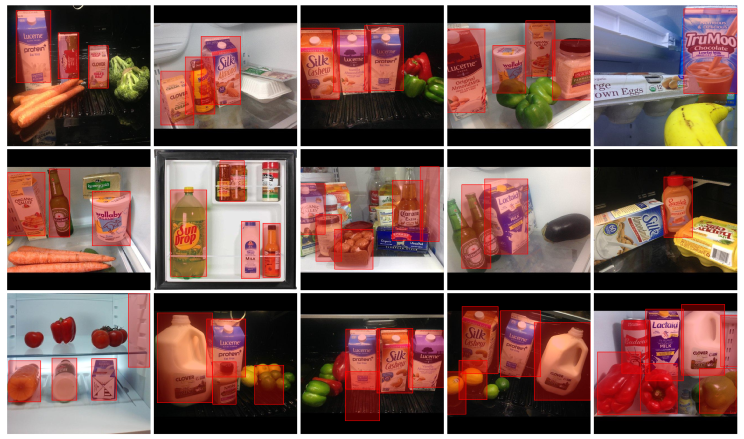}}
\end{center}
   \caption{Scenes representing variance in scale, background, textures, illumination, packing patterns and material properties wherein
   Top Row: Object detector correctly predicts the bound boxes for all objects of interest.
   Middle Row: Object detector misses objects of interest.
   Bottom Row: Object detector falsely predicts the presence of an object. }
\label{fig:PredictedInferences}
\end{figure}

We evaluate our best object detector model on a set of 50 crowd-sourced refrigerator scenes with all cue variances covering 55 distinct objects of interest considered as positives and 17 distractor objects as negatives. \autoref{fig:PredictedInferences} shows the variety in test set and the predicted bound-boxes for all refrigerator images. The detector achieves mAP of 24 on this dataset which is a promising result considering that no distractor objects were used while training using synthetic images.

We observe that detector handles scale, shape and texture variance. Though packing patterns like vertical stacking or highly oblique camera angles lead to false predictions. 
Few vegetables among the distractor objects are falsely predicted as objects of interest suggesting the influence of pre-training on ImageNet dataset also noting that the training dataset was devoid of such distractor objects marked as background clutter.
We report in \autoref{fig:TrainSizePlot},  \autoref{fig:LayerWisePlot} and \autoref{fig:DictSizePlot}   mAP vs. epochs trained plots over mAP vs. variance in factor to also represent the relevance of early stopping~\cite{Yao2007}. The networks trained by varying factors, show their peak performances for 25-50 epochs of training while the performance declines contrary to saturating which suggests over-fitting to synthetic images.

\begin{figure}[htbp]
\begin{center}
{\includegraphics[width=\linewidth]
{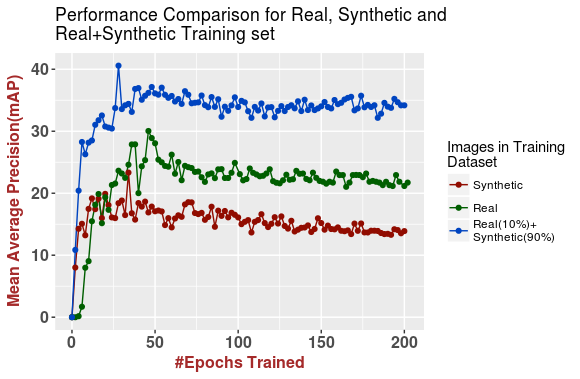}}
\end{center}
   \caption{Performance plots illustrating the effect of including  synthetic images while training neural networks.}
\label{fig:DetectionAccuracy}
\end{figure}
\section{Conclusion} \label{sec:conclusion}

The question arises how well does a network trained with synthetic images fare against
one trained with real world images. Hence we compare the performance of networks
trained with three different training image-sets as illustrated in
\autoref{fig:DetectionAccuracy}. The synthetic training set consisted of 4000 images
with 200 3D object models of interest while the real training set consisted of 400
images parsed from the internet with 240 distinct products and 19 distractor
objects. The hybrid set with synthetic and real images consisted of 3600
synthetic and 400 real images. All models were evaluated on a set of 50
refrigerator scenes with less than 5\% object overlap between  the test set and
train set images. CNN fully trained with 4000 synthetic images (achieves 24 mAP)
underperforms against one with 400 real images (achieves 28 mAP) but the addition
of 4000 synthetic images to real dataset boosts the detection performance by
12\% (achieves 36 mAP) which signifies the importance of transferable cues from
synthetic to real.

To improve the observed performance, several tactics can be tried. The presence
of distractor objects in the test set was observed to negatively impact
performance. We are working on the addition of distractor objects to the 3D
model repository for rendering scenes with distractor objects to train the network
to become aware of them. Optimizing the model architecture or replacing
DetectNet with object proposal networks might be another alternative. 
Training CNNs for semantic segmentation using synthetic images and the addition
of depth information to the training sets is also expected to help in the case
of images with high degree of occlusion.


\section*{Acknowledgment}
We acknowledge funding support from Innit Inc. consultancy grant CNS/INNIT/EE/P0210/1617/0007 and High Performance Computing Lab support from Mr. Sudeep Banerjee. We thank Aalok Gangopadhyay for the insightful discussions.



%\bibliographystyle{IEEEtran}
\bibliographystyle{IEEEtran} 
\bibliography{ref}

\end{document}